\documentclass[3p,11pt,nopreprintline]{elsarticle}
\usepackage{graphicx}
\usepackage{amsmath}
\usepackage{amssymb}
\usepackage{mathtools}
\usepackage{booktabs}
\usepackage{multirow}
\usepackage{makecell}
\usepackage[dvipsnames]{xcolor}
\usepackage{pifont}  
\usepackage{hyperref}
\usepackage{fontawesome}

\newcommand{\slow}{$^\text{\color{gray}\faHourglassStart}$}

\newcommand{\cmark}{\textcolor{Green}{\ding{51}}}
\newcommand{\xmark}{\textcolor{Red}{\ding{55}}}

\newcommand{\bx}{\boldsymbol{x}}
\newcommand{\bz}{\boldsymbol{z}}

\newcommand{\zloc}{z_1}
\newcommand{\zshape}{z_2}

\begin{document}

\begin{frontmatter}

\title{Odds-Structured Distributional Regression for Interpretable Time-to-Event Prediction}

\author[bifold,icm,charite]{Jakob Versnjak}
\author[bifold,ikim,charite]{Manuel Welte}
\author[icm,3,4,charite]{\mbox{Titus Kuehne}}
\author[icm,3,4,charite]{\mbox{Marcus Kelm}}
\author[bifold,ikim,charite]{\mbox{Grégoire Montavon}\corref{cor1}}

\cortext[cor1]{Corresponding author}

\address[bifold]{BIFOLD – Berlin Institute for the Foundations of Learning and Data, Berlin, Germany}
\address[icm]{Deutsches Herzzentrum der Charité, Institute of Computer-assisted Cardiovascular Medicine, Augustenburger Platz 1, 13353 Berlin, Germany}
\address[ikim]{Deutsches Herzzentrum der Charité, Institute for AI in Medicine, Invalidenstra{\ss}e 120/121, 10115 Berlin, Germany}
\address[3]{Department of Congenital Heart Disease -- Pediatric Cardiology, Deutsches Herzzentrum der Charité, Augustenburger Platz 1, Berlin 13353, Germany}
\address[4]{DZHK (German Centre for Cardiovascular Research), Partner Site Berlin, Potsdamer Str. 58, Berlin 10785, Germany}
\address[charite]{Charité -- Universitätsmedizin Berlin, Corporate Member of Freie Universität Berlin and Humboldt-Universität zu Berlin, Charitéplatz 1, Berlin 10117, Germany}

\begin{abstract}
Time-to-event prediction is fundamentally challenged by partial observability and aleatoric uncertainty. In clinical medicine, for instance, the timing of disease progression or acute readmission is contingent on highly dynamic and largely latent health trajectories. Rather than forcing a deterministic point forecast, distributional regression offers a principled means to address this complexity by predicting a probability distribution over time. However, explaining these distributional predictions presents a key obstacle: raw distribution parameters output by the model are abstract mathematical quantities that do not align well with domain concepts.
To gain interpretability, we propose a novel paradigm where output parameters are designed to be additive in log-odds---a widely established measure in clinical risk modeling. Our approach, called ODDS-DR, moves beyond technical convenience to deliver clear domain alignment as well as the layer-wise computational structure required for robust, time-dependent feature attribution.
Validated across two real-world clinical cohorts, our framework matches state-of-the-art predictive performance while resolving distributional outputs into interpretable risk markers across flexible prognostic horizons.
\end{abstract}


\end{frontmatter}

\section{Introduction}

In real-world applications---and particularly within clinical medicine---predicting time-to-event outcomes is fundamentally challenged by latent physiological states, sparse monitoring schedules, and the inherent variability of downstream clinical actions and care pathways. Consequently, critical endpoints---such as acute readmission, disease progression, therapeutic response, or mortality---exhibit profound temporal stochasticity. Capturing the full complexity of temporal uncertainty requires more than summarizing it via error bars or rigid hazard ratios; it demands modeling the event's full probabilistic distribution over time.

\medskip

Distributional regression \cite{bishop94, Stasinopoulos2007, Rasp2018} provides the flexibility to capture this temporal complexity. Unlike classical Cox-based semi-parametric estimators \cite{Cox1972,katzman2018deepsurv}---which enforce time-invariant proportional hazard ratios---distributional regression parameterizes the complete conditional density function via distinct location, scale, and shape parameters. In practice, however, outputting an accurate predictive distribution is rarely sufficient. Domain experts value the factors driving a prediction as much as the forecast itself. This emphasis on interpretability is ubiquitous in clinical fields, where identifying actionable risk markers is often as critical as the risk estimate itself \cite{Lundberg2020, Keyl2025, FIERLEY2026104041}. In time-to-event tasks, this means looking beyond event timing alone to disentangle which factors operate across different prognostic horizons---separating long-term baseline predispositions (e.g.,\ genetics) from short-term metabolic or lifestyle disruptions.

\medskip

However, a key methodological obstacle limits the practical utility of distributional regression models. The parameters produced at their output are inherently technical and specific to the chosen distribution family. Identifying features that drive these raw parameters risks severe misalignment with domain experts. Model-agnostic explainers \cite{Strumbelj2013, Lundberg2020} can in principle access and attribute interpretable readouts from the predicted distributions \emph{post-hoc}, but their sampling-based mechanisms are computationally demanding and brittle in the presence of nonlinearities.

\medskip

\begin{figure}[t!]
    \centering
    \makebox[\textwidth][c]{\includegraphics[width=1\textwidth]{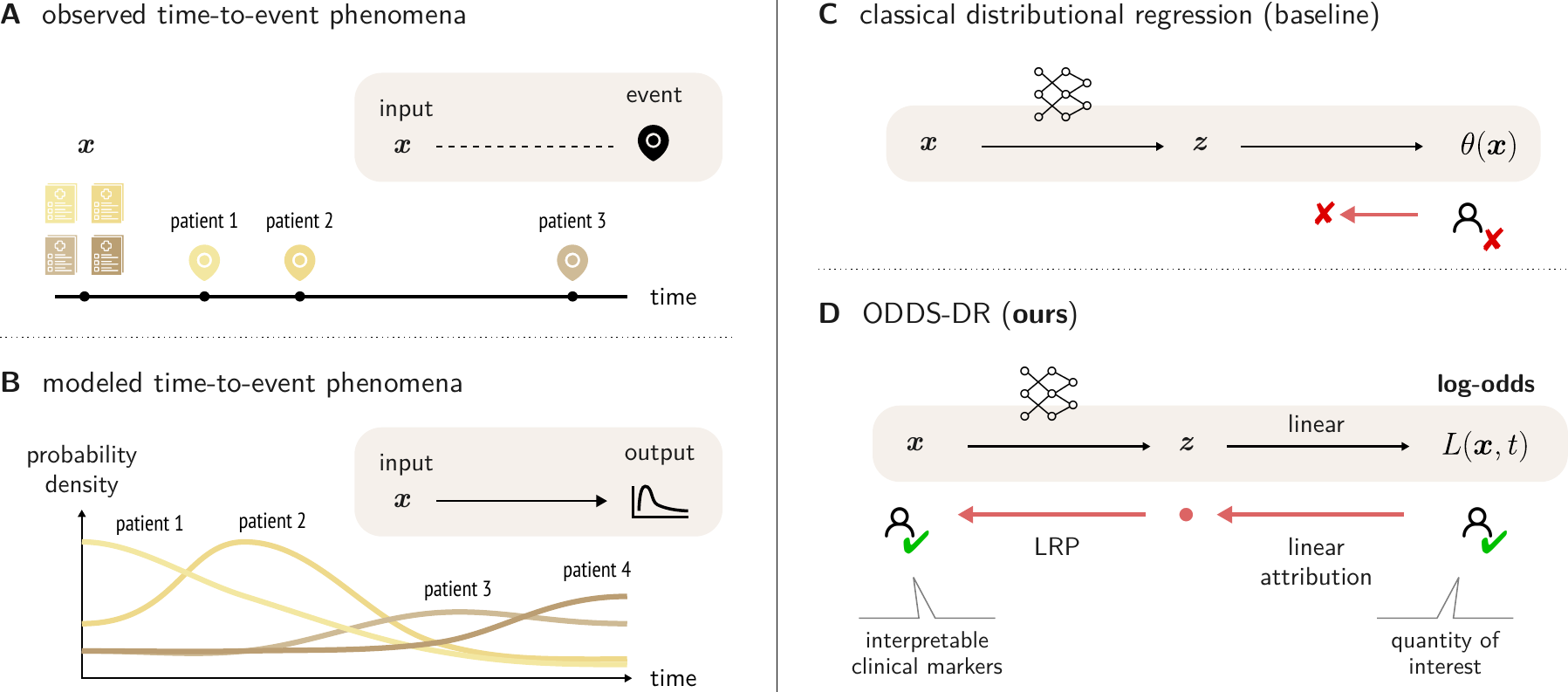}}
\caption{{Time-to-event modeling and proposed ODDS-DR framework.} 
\textbf{(A)}~Clinical data records both the occurrence of events and their temporal organization.
\textbf{(B)}~Distributional regression models this temporal dynamic by parameterizing continuous, patient-specific probability densities over time.
\textbf{(C)}~{Classical baseline:}~Standard distributional regression predicts distribution-specific parameters $\theta(\boldsymbol{x})$ (e.g., shape and scale) that lack clinical interpretability and decomposability into input features. 
\textbf{(D)}~{ODDS-DR (ours):} By parameterizing model representations to contribute linearly to event log-odds $L(\boldsymbol{x}, t)$---a clinically established quantity---our framework enables exact linear attribution to intermediate representations, followed by propagation (e.g., via LRP) back to input clinical features.}
    \label{fig:diagram}
\end{figure}

In this paper, we address these limitations by redesigning distributional regression models to natively embed log-odds interpretability. Unlike bounded probabilities or family-specific distribution parameters, the log-odds maps evidence additively onto the real line, while unifying diverse distribution families under a single, comparable metric. For an event occurring at time $T$, the cumulative log-odds by time $t$ is defined as:
\begin{equation}
\label{eq:log-odds-intro}
L(t) = \log \left( \frac{P(T \leq t)}{P(T > t)}\right)\,.
\end{equation}
We propose a novel reparameterization where the network outputs combine additively to reconstruct $L(t)$ across arbitrary horizons $t$. We derive this additive structure for Gaussian, Logistic, and Log-Logistic families. In each case, it yields the layer-wise computational graph required for fast, robust propagation-based attribution methods such as Layer-wise Relevance Propagation (LRP) \cite{Bach2015, Montavon2019}, where explanations at any time horizon can be retrieved as a linear combination of only two explanation prototypes. Consequently, practitioners can query specific temporal horizons (e.g., ``what factors drive this patient's risk of surgical reintervention within six months?'') or track risk markers across time (e.g., ``how does the TISS score influence mortality risk from the acute window into the long term?'') consistently and orders of magnitude faster than with model-agnostic explainers. Our framework, termed \emph{Odds-Structured Distributional Regression} (ODDS-DR), is illustrated in Fig.~\ref{fig:diagram}.

\medskip

We validate our approach across two real-world clinical cohorts: the public SUPPORT2 survival dataset and an institutional time-to-reintervention cohort of aortic coarctation (CoA). In both cases, our framework matches or exceeds the predictive performance of standard baselines (e.g., DeepSurv \cite{katzman2018deepsurv}, DeepHit \cite{deephit}, Deep Survival Machines \cite{nagpal2021deep}) while uncovering risk factors that closely align with established clinical guidelines. Beyond achieving clinical alignment on conventional time ranges, we demonstrate our method's capability to extend evidence-based risk assessment to arbitrary temporal horizons and individual patient profiles.

\section{Related work}

Rigorously modeling, quantifying, and explaining time-to-event uncertainty intersects three active areas of literature. In Section~\ref{sec:related-distributional}, we review continuous distributional regression and heteroskedastic modeling. In Section~\ref{sec:related-survival}, we contrast continuous parametric density formulations against conventional semi-parametric and discrete time-to-event paradigms. Finally, in Section~\ref{sec:related-xai}, we examine existing feature attribution methods for uncertainty and time-to-event predictions. The advantages and limitations of the methods described below are summarized in Table~\ref{tab:model_comparison}.

\subsection{Heteroskedastic Modeling and Distributional Regression}
\label{sec:related-distributional}

A first simple class of models that embeds uncertainty directly within the model structure is heteroskedastic regression, which adds to each model output a predictive spread that depends on the input \cite{Nix1994}. Another class of models, distributional regression---formalized via Generalized Additive Models for Location, Scale, and Shape (GAMLSS) \cite{Stasinopoulos2007} and Mixture Density Networks \cite{bishop94}---generalizes this principle by simultaneously learning distinct functional dependencies for location, scale, and shape parameters to yield complete conditional densities. Modern semi-structured extensions integrate deep neural networks with structured additive predictors \cite{rügamer2022semistructureddistributionalregression}, enabling high-capacity representation learning while retaining transparent, parameter-specific additive components. In survival analysis, Deep Conditional Transformation Models (DCTMs) \cite{dctm, dctm-survival-MICCAI2025} reparameterize distributions via Bernstein polynomials, which yields an additive decomposition of log-odds for the logistic distribution.

In contrast to these works, our ODDS-DR framework establishes the design of parameterizations that are log-odds-additive as its core principle, applicable to a broad range of distributions. To the best of our knowledge, ODDS-DR is also the first framework that exploits log-odds additive structures both for prediction and feature attribution.

\subsection{Time-to-Event Frameworks}
\label{sec:related-survival}

A significant body of work addresses uncertain outcomes specifically in time-to-event analysis, where Cox Proportional Hazards (PH) models \cite{Cox1972} form the dominant paradigm. DeepSurv \cite{katzman2018deepsurv} provides a modern extension of this framework, parameterizing the log-hazard with a neural network to capture non-linear feature interactions, while retaining the original assumption of time-invariant relative hazards. Cox-Time \cite{KvammeBS19} relaxes this constraint by providing time as an additional network input, allowing relative hazards to depend jointly on features and time. Because the network outputs an instantaneous hazard rate, obtaining survival curves at test time requires numerically integrating these rates over time. To keep this computation tractable, predictions are evaluated on a discretized time grid derived from a subset of the training data. Specifically designed for interpretability, architectures such as Cox-SE \cite{Alabdallah2026} and CoxNAM \cite{XU2023120218} compute sample-specific or additive feature contributions directly within the network while retaining the proportional hazards formulation. Departing from Cox-based formulations, another line of work partitions the continuous time axis into discrete intervals, using architectures such as DeepHit \cite{deephit} or neural multi-task formulations \cite{gensheimer2019nnetsurvival, kvamme2021continuous} to model non-linear, time-varying event probabilities across pre-specified bins.

More closely related to our ODDS-DR framework, prior work addresses time-to-event modeling through distributional regression \cite{weibull_survival, bennis2020estimationconditionalmixtureweibull, dctm-survival-MICCAI2025}. Within this family of methods, Deep Survival Machines (DSM) \cite{nagpal2021deep} allows for the expression of fine-grained time-to-event profiles by letting neural networks output complex parametric forms (such as mixtures of Weibull or Log-Normal distributions). In contrast to Cox-based approaches, the distributional regression framework provides a variety of advantages: the likelihood factorizes into independent per-sample contributions, enabling direct applicability of fast mini-batch-based training. Furthermore, direct access to the predicted distribution at the output simplifies multi-horizon querying and attribution.

Our proposed framework, ODDS-DR, inherits the key advantages of distributional regression, particularly its expressivity and integration into neural network training pipelines. Crucially, what distinguishes ODDS-DR from existing distributional regression methods is its focus on native explainability. Specifically, by parameterizing predicted distributions to be additive in log-odds, our framework yields feature attributions that are both inherently interpretable and structured across entire time horizons.

\begin{table*}[t!]
\centering \small
\makebox[\textwidth][c]{
\setlength{\tabcolsep}{4pt}
\begin{tabular}{lcccccccc}
\toprule
\textbf{Property} 
& \makecell{Classifier} 
& \makecell{Cox\\\cite{Cox1972}}
& \makecell{DeepSurv\\\cite{katzman2018deepsurv}} 
& \makecell{Cox-Time\\\cite{KvammeBS19}}
& \makecell{DeepHit\\\cite{deephit}}
& \makecell{DSM\\\cite{nagpal2021deep}}
& \makecell{CoxNAM\\\cite{XU2023120218}}
& \makecell{ODDS-DR\\(\textbf{ours})} \\
\midrule

\textit{Modeling capability}\\

~~~Continuous event time 
& \xmark & \cmark & \cmark & \cmark & \xmark$^\text{(b)}$ & \cmark & \cmark & \cmark \\

~~~Right-censoring support 
& \xmark & \cmark & \cmark & \cmark & \cmark & \cmark & \cmark & \cmark \\

~~~Time-varying effects 
& \xmark & \xmark & \xmark & \cmark & \cmark & \cmark & \xmark & \cmark \\

~~~Non-linear representations 
& \cmark & \xmark & \cmark & \cmark & \cmark & \cmark & \cmark & \cmark \\[2mm]

\textit{Interpretability}\\

~~~Multi-horizon attribution 
& \xmark & \xmark & \xmark & \cmark & \cmark & \cmark & \xmark & \cmark \\

~~~Time-parametric attribution$^\text{(a)}$ 
& \xmark & \xmark & \xmark & \xmark & \xmark & \xmark & \xmark & \cmark \\[2mm]

\textit{Optimization \& inference}\\

~~~Mini-batch loss decomposition 
& \cmark & \xmark & \xmark & \cmark & \cmark & \cmark & \xmark & \cmark \\

~~~Closed-form risk inference 
& \cmark & \cmark & \cmark & \xmark$^\text{(c)}$ & \cmark & \cmark & \cmark & \cmark \\
\bottomrule
\end{tabular}
}
\caption{{Systematic comparison of ODDS-DR with existing time-to-event frameworks.} 
Symbols: \cmark{} supported natively; \xmark{} not supported. 
Superscripts denote methodological specifics: 
$^\text{(a)}$\,Attributions are continuous, closed-form functions of time (see definition in Section~\ref{section:odds_dr}); 
$^\text{(b)}$\,Relies on a discretized time grid rather than a continuous-time formulation; 
$^\text{(c)}$\,Requires numerical integration over a discrete grid rather than an analytical closed-form evaluation.}
\label{tab:model_comparison}
\end{table*}

\subsection{Explainable AI Frameworks for Probabilistic Outcomes}
\label{sec:related-xai}

While Explainable AI frameworks are well established for scalar point predictions \cite{Strumbelj2013, Bach2015, Lundberg2017APredictions}, fewer works have addressed the question of how to attribute continuous probabilistic outcomes. Some recent efforts focus primarily on explaining dispersion statistics, such as ensemble variance \cite{BLEY2025111171} or heteroskedastic Gaussian spread \cite{iversen2025identifyingdriverspredictivealeatoric}. A series of recent works have addressed the explainability of probability-linked quantities as they occur in time-to-event models. Specific to Cox-based models, Cox-SE \cite{Alabdallah2026} and CoxNAM \cite{XU2023120218} parameterize instance-level feature contributions within the log-hazard ratio to facilitate the explanation of this quantity.

Leveraging post-hoc interpretability methods that can readily apply to any black-box model---including distributional regression---the authors of \cite{langbein2025gradientbasedexplanationsdeeplearning} adapt classical gradient-based explanation techniques to survival, enabling explanations resolved across both input features and time. Perturbation-based estimators have similarly been extended over time: SurvSHAP(t) \cite{Krzyzi_ski_2023} computes time-resolved Shapley values across the prediction horizon, SurvLIME(t) \cite{kovalev2020survlimemethodexplainingmachine} fits local Cox surrogates, and SurvBeNIM \cite{Utkin2025SurvBeNIMTB} constructs non-parametric local survival approximations. More recently, SurvFD and SurvSHAP-IQ \cite{langbein2026functionaldecompositionshapleyinteractions} decompose time-dependent and time-independent feature interactions.

ODDS-DR departs from these existing frameworks by addressing explainability natively within the model architecture, specifically by outputting parametric forms that are additive in the log-odds. This design enables the application of propagation-based attribution methods such as LRP \cite{Bach2015, Montavon2019}, which are significantly faster to compute and more robust than post-hoc methods reliant on repeated sampling of nonlinear functions. In particular, ODDS-DR's structure allows generating explanations resolved across an entire time horizon from only two prototypical attributions.

\section{Odds-Structured Distributional Regression (ODDS-DR)}
\label{section:odds_dr}

Standard neural distributional regression predicts the parameters of a continuous conditional density $p(T\,|\,\bx)$---such as location and scale $(\mu, \sigma)$ for Gaussian or Logistic models, or $(\alpha, \beta)$ for Log-logistic distributions. However, aligning network outputs with these raw mathematical parameters is largely a technical convenience that lacks principled grounding in either modeling or interpretability. Our proposed method, ODDS-DR, introduces a novel paradigm where the model outputs additively predict cumulative log-odds across continuous prognostic horizons.

\smallskip

Let $\bx \in \mathcal{X}$ denote an input feature vector, $T \in \mathbb{R}^+$ a continuous target variable measuring time to event, and $t$ a prognostic horizon. Our framework designs the model outputs $z_j(\bx) \in \mathbb{R}$ so that they decompose the cumulative log-odds as:
\begin{align}
L(\bx,t) &= \log \left( \frac{P(T \leq t \mid \bx)}{P(T > t \mid \bx)} \right)
\label{eq:logodds-def}\\
&= \sum_j \lambda_j(t)\, z_j(\bx),
\label{eq:logodds-linear}
\end{align}
where the weighting coefficients $\lambda_j(t)$ depend solely on the horizon $t$. This linear parameterization of ODDS-DR offers two key advantages over classical parameterizations and broader time-to-event methods.

\begin{description}
\item[Property 1: Epistemic Inductive Bias.] The homogeneous linear relation between model outputs and log-odds ensures that uncertainty is naturally preserved in two fundamental modes: feature missingness and out-of-domain prediction. In both regimes, a properly regularized model tends to produce outputs close to zero (e.g., owing to the positive homogeneous tendency of neural networks or to attenuated responses outside the data support). These behaviors yield the natural implication chains:
\begin{align*}
\bx \approx \boldsymbol{0} &\implies \bz \approx \boldsymbol{0} \implies L \approx 0 \quad \text{(feature missingness)}, \\
p(\bx) \approx 0 &\implies \bz \approx \boldsymbol{0} \implies L \approx 0 \quad \text{(data outlierness)}.
\end{align*}
This conservative inductive bias is essential in high-stakes domains---such as clinical risk stratification on underrepresented patient subgroups---as it prevents the model from generating spurious, unsupported certainty.

\item[Property 2: Fast Time-Parametric Attribution.] Leveraging the common linearity axiom of explanations, attributing the log-odds to the input features can be reduced to explaining individual output neurons:
\begin{align}
\mathcal{E}\Big\{\sum_j \lambda_j(t)\, z_j(\bx)\Big\} = \sum_j \lambda_j(t) \, \mathcal{E} \big\{z_j(\bx)\big\}.
\label{eq:weighted-expl}
\end{align}
Specifically, calculating $\mathcal{E} \big\{z_j(\bx)\big\}$ is a staple in Explainable AI, with methods such as LRP requiring only one backward pass. Consequently, attributing log-odds across an entire continuum of prognostic horizons demands only as many backward passes as there are output neurons---two for the two-parameter distributions considered in this work---which is orders of magnitude more efficient than model-agnostic approaches. Furthermore, Eq.~\eqref{eq:weighted-expl} reveals an intrinsic low-rank structure in multi-horizon attribution, providing a compact basis to analyze how risk markers vary across prognostic horizons.
\end{description}
In the following subsections, we derive the explicit mappings relating network outputs, log-odds, and the standard parameters of common distribution families used to predict events under diverse uncertainty profiles.

\subsection{ODDS-DR for Low-Uncertainty Events}
\label{section:reparameterization-bell}

\begin{figure}[t!]
    \centering
    \makebox[\textwidth][c]{%
    \includegraphics[width=1\linewidth]{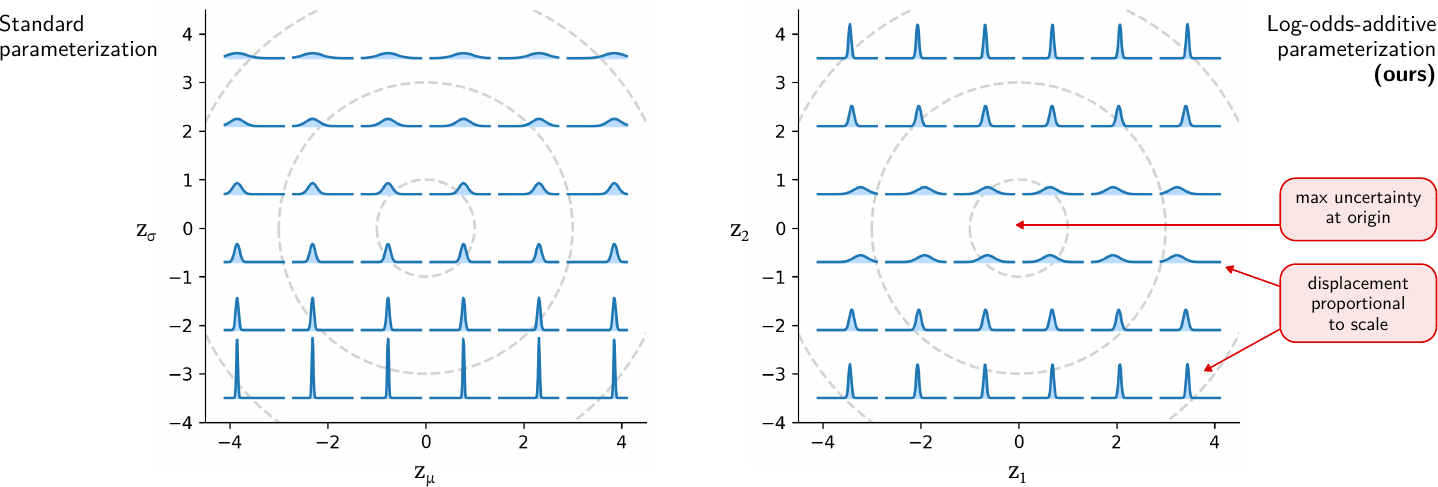}
    }
    \caption{{Comparison of parameterization geometries for a Gaussian distribution.} 
{Left (Standard parameterization):} Classical formulation with location $\mu = z_\mu$ and scale $\sigma = \operatorname{softplus}(z_\sigma)$, where variance changes monotonically along the vertical axis independently of location. 
{Right (ODDS-DR, log-odds-additive parameterization):} Our reparameterization $(z_1, z_2)$ places maximum uncertainty (spread) at the origin $(0,0)$, with lateral displacement along $z_1$ scaling proportionally with density spread. This geometry guarantees exact log-odds additivity and facilitates feature attribution.}
    \label{fig:radar}
\end{figure}

Regularly recurring physiological events (e.g., cardiac cycles, respiratory intervals, or sensory evoked responses), where deviations from a baseline rhythm indicate underlying pathology, can be categorized as low-uncertainty events. Tracking the log-odds of event occurrence relative to a reference duration holds clinical value in deciding whether an event is pathologically delayed or imminent. Bell-shaped distributions---including Gaussian, Laplace, or Logistic distributions---are a natural model of deviations around an overall predictable schedule. The log-odds of such distributions can be expressed as $L = g((t-\mu)/\sigma)$, where $g$ is typically a monotonically increasing odd function. For the Gaussian case, it is given by $g(\xi) = 2 \tanh^{-1}(\operatorname{erf}(\xi/\sqrt{2}))$, whereas for the heavier-tailed Logistic distribution, it is simply the identity function $g(\xi) = \xi$ (derivations in \ref{appendix:log-odds-bell-shaped}). This expression can be Taylor-expanded in the vicinity of some baseline $t_0$ as:
\begin{align}
L(\bx,t) &\approx \underbrace{g \bigg(\frac{t_0-\mu}{\sigma}\bigg)}_{z_1(\bx)}
\,\cdot\,
\underbrace{\bigg.1}_{\lambda_1(t)}
\,+\,
g' \bigg(\frac{t_0-\mu}{\sigma}\bigg) 
\cdot
\underbrace{\bigg(\frac{1}{\sigma}\bigg)}_{z_2(\bx)}
\,\cdot\,
\underbrace{\bigg.(t - t_0)}_{\lambda_2(t)} \label{eq:logodds-bell-taylor}
\end{align}
where we recover a structure similar---or equivalent in the Logistic case ($g'=1$)---to that of Eq.\ \eqref{eq:logodds-linear}, with separate input-dependent and horizon-dependent terms. The motivation for not absorbing $g'$ in the output $z_2(\bx)$ is to ensure a closed-form relation between outputs and model parameters:
\begin{align*}
    \sigma &= 1 / z_2,\\
    \mu &= t_0 - \frac{g^{-1}(z_1)}{z_2},
\end{align*}
which is convenient for model training. Lastly, to enforce the positivity of $\sigma$, we define $z_2 = \|\boldsymbol{z}_2\|_2$ as the Euclidean norm over a vector of pre-outputs. While this serves a similar role to a standard $\operatorname{softplus}$ activation, it preserves positive homogeneity, which is desirable for fulfilling \textbf{Property 1} as well as stable propagation-based attribution via LRP. The relation between the outputs and the probability density function is shown in Fig.\ \ref{fig:radar}.

\subsection{ODDS-DR for High-Uncertainty Events}
\label{section:reparameterization-lifetime}

While symmetric, bell-shaped location-scale distributions are suited for deviations around a recurring rhythm, they may fall short for events with high temporal uncertainty, such as disease onset, mortality, and need for reintervention. Lifetime distributions---including log-normal, log-logistic, and Weibull forms---provide a highly diffuse and flexible alternative by stretching probability mass over a much broader timeline, allowing it to both concentrate near $t=0$ and extend out to very large time values. Focusing specifically on the Log-logistic distribution with parameters $(\alpha,\beta)$ and proceeding similarly as in Section \ref{section:reparameterization-bell}, we can express the log-odds additively as a function of $\log(t)$:
\begin{align}
L(\bx,t)
&= \beta \cdot (\log(t) - \log(\alpha))
\label{eq:logodds-lifetime}\\
&=
  {\underbrace{\beta \cdot (\log(t_0) - \log(\alpha))}_{z_1(\bx)}} \cdot \underbrace{1}_{\lambda_1(t)}
+ 
{\underbrace{\beta}_{z_2(\bx)}} \cdot \underbrace{(\log(t) - \log(t_0))}_{\lambda_2(t)}, \label{eq:logodds-lifetime-taylor}
\end{align}
which forms an exact linear decomposition centered at an operating horizon $\log(t_0)$. The derivation is given in \ref{appendix:log-odds-log-logistic}.
This expansion again leads to the identification of two additive terms, where outputs $z_1$ and $z_2$ can be defined to implement the generic form of Eq.\ \eqref{eq:logodds-linear}. Rearranging the equation above, we get the following expressions for the Log-Logistic parameters:
\begin{align*}
\beta &= \zshape,\\
\alpha &= \exp(- \zloc / \zshape) \cdot t_0.
\end{align*}
To enforce the positivity requirement of $z_2$, we proceed as in Section \ref{section:reparameterization-bell} by defining it as the norm of a vector of pre-outputs. An analysis of the reparameterization above yields a similar interpretation as for the Bell-shaped cases: Output neurons with value close to zero produce a near-zero shape parameter $\beta$, corresponding to a maximum level of uncertainty (log-odds of zero).

\subsection{Accounting for Event Non-Occurrence}

In the presence of competing events or non-susceptibility, the primary event may never occur (e.g., due to acquired immunity or lifetime avoidance). Such event non-occurrence can be formalized via a mixture model where $\eta \in (0, 1)$ denotes the marginal probability of ever experiencing the event, and $1-\eta$ denotes the non-occurrence probability. Under this mixture, the log-odds of the event occurring before time $t$ can be expressed as:
\begin{align}
L^{\mathrm{(new)}} &= \log \left(\frac{\eta \cdot P(T \leq t)}{\eta \cdot P(T > t) + (1-\eta)} \right)
\label{eq:competing-1}\\
&= \mathrm{smin}\left(L + \log \eta \,,\, L_0\right),
\label{eq:competing-2}
\end{align}
where $L$ denotes the baseline log-odds conditional on event occurrence (cf.\ Eq.~\eqref{eq:logodds-def}), $L_0 = \log [\eta / (1-\eta)]$ is the marginal log-odds of susceptibility, and $\mathrm{smin}(a, b) = -\log(e^{-a} + e^{-b})$ is the smooth minimum. 

Equation~\eqref{eq:competing-2} offers an intuitive asymptotic interpretation. On a \textit{short-term horizon} ($t \to 0$, $L \ll 0$), the first term dominates, yielding the linear approximation $L^{\mathrm{(new)}} \approx L + \log \eta$. Thus, short-term risk remains governed by the original outputs $z_1$ and $z_2$, shifted by the base rate $\log \eta$. Conversely, on a \textit{long-term horizon} ($t \to \infty$, $L \gg 0$), the second term dominates such that $L^{\mathrm{(new)}} \to L_0$. The predicted log-odds naturally plateaus at the marginal log-odds of susceptibility, preventing runaway over-prediction.

\section{Qualitative Experiments on Synthetic Data}
\label{section:toy}

\begin{figure}[t!]
    \centering
    \makebox[\textwidth][c]{%
    \includegraphics[width=0.925\linewidth]{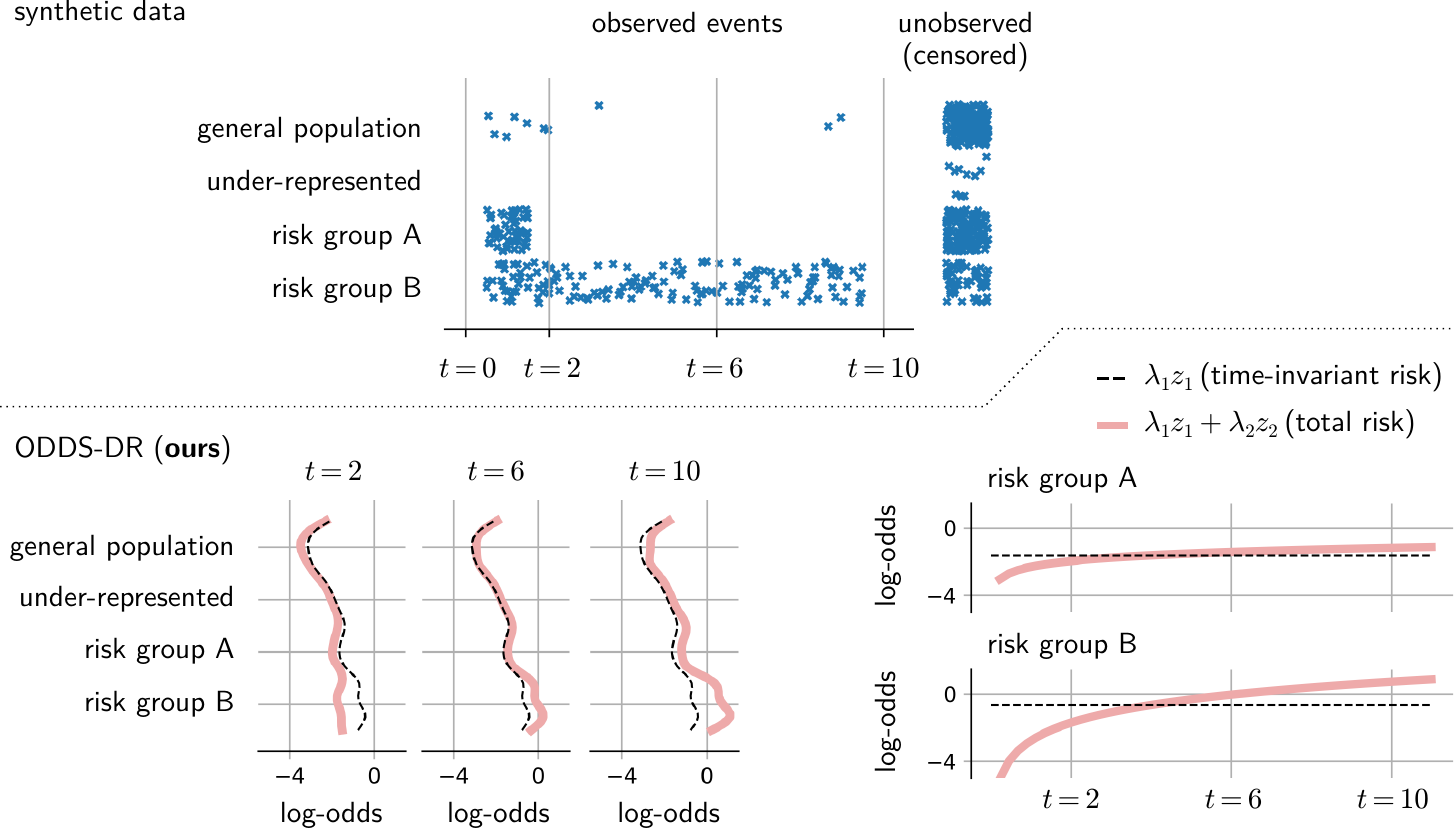}
    }
\caption{{ODDS-DR behavior on synthetic time-to-event data with heterogeneous risk profiles and representativeness gaps.} 
{Top:} Observed and censored event times across four subpopulations: general population, under-represented, risk group~A (events concentrated early at $t \le 2$), and risk group~B (events distributed over an extended horizon $t \le 10$). 
{Bottom left:} Predicted log-odds across subpopulations evaluated at horizons $t \in \{2, 6, 10\}$, decomposing total risk ($\lambda_1 z_1 + \lambda_2 z_2$, pink) into time-invariant baseline risk ($\lambda_1 z_1$, dashed) and time-dependent adjustment. 
{Bottom right:} Continuous risk trajectories over time for risk groups~A and B, demonstrating how ODDS-DR captures distinct temporal risk dynamics.}
    \label{fig:toy}
\end{figure}

We build a toy dataset that synthesizes some main aspects of timing clinical events. First, a phenotype representation $\bx$ of the individual, here embedded in a one-dimensional input space. Second, the time $T$ of event occurrence, which is also one-dimensional, with a limited 10-month observational period. The input and time-to-event targets form a joint distribution that can be depicted as a two-dimensional scatter plot, shown in Fig.\ \ref{fig:toy} (top). The dataset is designed to exhibit the following joint input-target structure: \textbf{(i)} a \textit{general population} that rarely experiences the clinical event within the 10-month observational time frame, \textbf{(ii)} an \textit{underrepresented} subgroup that appears to share the statistics of the general population but that is distinctly undersampled, \textbf{(iii)} \textit{risk group A}, which occasionally experiences the clinical event early after baseline, emulating short-term risks, and \textbf{(iv)} \textit{risk group B}, experiencing the clinical event more frequently but over a longer time period, emulating longer-term risks.

\smallskip

We train a log-logistic ODDS-DR model on this data. From the one-dimensional input, it produces the outputs $z_1 \in \mathbb{R}$ and $z_2 = \|\bz_2\|$ with $\bz_2 \in \mathbb{R}^3$. The mapping from input to outputs is achieved by a two-layer model, specifically, a static Gaussian feature expansion layer followed by a linear layer. The model is trained to maximize log-likelihood (full-batch, 2500 epochs, momentum of 0.85, learning rate: 0.01, weight decay: 0.01). The log-odds profile $L(\bx,t)$ produced by our model after training can be navigated both in the input domain and across time, and further decomposed using Eq.\ \eqref{eq:logodds-linear} into two terms: a largely time-invariant baseline $\lambda_1(t) z_1(\bx)$, and another term $\lambda_2(t) z_2(\bx)$ capturing the time-bound fluctuations. This analysis is depicted in Fig.\ \ref{fig:toy} (bottom).

\smallskip

As expected, the general population is predicted to have significantly lower log-odds than the risk subgroups. Distinctly, our method demonstrates the capability to assign higher risk to the underrepresented subgroup in spite of an absence of observed clinical events in the data. This advantageous behavior derives directly from ODDS-DR's \textbf{Property 1} presented in Section \ref{section:odds_dr}, where low output activations arising from missing evidence or lack of support drive the parametric family to its mode of high uncertainty.

\smallskip

Looking more closely at the two risk groups, we observe that they have comparable time-independent risks (captured by $\lambda_1(t) z_1(\bx)$ and displayed as a dotted line), but differ significantly with respect to the horizon-dependent contributor $\lambda_2(t) z_2(\bx)$, causing overall risk to be similar on a short-term horizon ($t=2$) while steadily growing for risk group B over the longer term ($t=6$ and $t=10$). This ability of our method to represent risk profiles that are both subgroup-specific and time-dependent is also observed in longitudinal profiles (Fig.\ \ref{fig:toy}, bottom right), where both risk profiles follow a parametric structure (\textbf{Property~2} presented in Section~\ref{section:odds_dr}), with risk group~B exhibiting a noticeably steeper progression.

\section{Experiments on Real-world Clinical Cohorts}
\label{section:realworld}

Broadening our analysis to real-world multivariate data, we evaluate our method on two clinical cohorts featuring varying population sizes, types of events, multi-dimensional panels of clinical measurements, as well as missing values and differing levels of right-censoring.

\smallskip

The first dataset, SUPPORT2 \cite{knaus2020support}, derives from the `Study to Understand Prognoses and Preferences for Outcomes and Risks of Treatment'---a prospective multi-centre observational study including 9,105 severely ill patients across five US clinical centres, recruited during 1989--1991 and 1992--1994. The primary outcome of interest was all-cause mortality. Data include 67 different features covering demographics, clinical and laboratory measurements, disease characteristics, as well as clinician- and model-based prognostic estimates. The median observed event time was 58 days (IQR: 14--252 days), with 32\% right-censoring. Missingness varied across features, reaching up to 53\% for laboratory measurements. With its large sample size, this dataset is ideal for testing the expressive power of time-to-event models. Although SUPPORT2 reflects clinical practice from more than 30 years ago, it remains a widely used benchmark in survival analysis and Explainable AI, enabling methodological comparison. We therefore use it primarily for methodological validation of ODDS-DR and acknowledge that the clinical relevance of individual risk markers may not fully transfer to contemporary practice.

\smallskip

The second dataset, CoA, is a prospective, well-characterized single-centre observational study conducted at the Deutsches Herzzentrum der Charité, comprising 160 patients with aortic coarctation (CoA) across 218 visits recruited between 2014 and 2022 \cite{FIERLEY2026104041}. The cohort provides a clinically detailed and disease-specific setting for evaluating the proposed framework, combining 37 structured clinical features extracted from routinely collected electronic health records, echocardiography and cardiovascular magnetic resonance (CMR) imaging. Missing values are present across imaging-derived features, generally below 35\%, except for two CMR parameters, which are missing in 51\% of visits. The event of interest was re-coarctation requiring invasive reintervention, either catheter-based (stenting or balloon angioplasty) or surgical. The median observed event time was 21.5 months (IQR: 8.5--41.25 months). The cohort is inherently limited in size due to the low prevalence of the disease and further characterized by high right-censoring (75\%), posing distinct regularization challenges.

\smallskip

Both datasets used a 70/30 training-validation split. SUPPORT2 was stratified by event status, whereas CoA used the same split as in our recent publication \cite{FIERLEY2026104041}, stratified by sex, age, and event status.
Overall, these two datasets provide broad coverage, spanning large-scale, low-noise conditions to intrinsically smaller-scale, high-noise settings. Their associated clinical task also comes in each case with clinical guidelines that we use as ground-truth signals to evaluate not only prediction accuracy, but also the plausibility of extracted risk markers.

\subsection{Training and Hyperparameter Selection} \label{sec:model_comparison}

In our experiments, outputs $z_1 \in \mathbb{R}$ and $\boldsymbol{z}_2 \in \mathbb{R}^3$ are predicted by a fully-connected feed-forward neural network with 2--4 hidden layers and 16--256 neurons per layer, selected through hyperparameter tuning. For the CoA cohort, dropout was fixed at 0.5 to improve regularization and training stability, whereas for SUPPORT2 it was optimized over 0--0.5. Each hidden layer is followed by a ReLU activation. Features were standardized to zero mean and unit variance, with normalization statistics estimated exclusively from the observed (non-missing) values of each feature $x_i$, such that $\mathbb{E}[x_i]=0$ and $\mathrm{std}(x_i)=1$. To account for missing values in the input data, each standardized feature $x_i$ was transformed into a new input representation via the map
\begin{equation}
x_i \mapsto (1-x_i,\,1+x_i) \cdot p_i
\end{equation}
where $p_i \in \{0,1\}$ denotes feature presence, with $p_i=1$ if observed and $p_i=0$ if missing. This reserves the zero-input to feature missingness, distinguishing it from an actual observed zero value. This promotes the default mode of full feature missingness mapping to the zero output vector, enabling us to fulfill ODDS-DR's \textbf{Property 1} (cf. Section \ref{section:odds_dr}) in an end-to-end fashion. The model was trained using stochastic gradient descent (learning rate: $10^{-4}$--$10^{-2}$; momentum: 0--0.95; weight decay: $10^{-8}$--$10^{-2}$) while accounting for right-censored observations in the loss function. 

All preprocessing, including feature scaling, was performed exclusively in the training cohort and subsequently applied to the validation cohort. Hyperparameters were optimized within the training cohort using cross-validation and Bayesian optimization.

\subsection{Baselines}

Our model was evaluated against a representative set of methods used for survival analysis, including classical distributional regression with classical parameterizations, the well-established Cox proportional hazard model \cite{Cox1972} (\ref{appendix:cox}), deep extensions of the latter (DeepSurv \cite{katzman2018deepsurv} and Cox-Time \cite{KvammeBS19}), as well as the non-parametric DeepHit model \cite{deephit}. The same neural network architecture was used for all distributional models, differing only in the output layer: in the baseline models, the outputs directly represented the distributional parameters, whereas our parameterization made the outputs linearly related to the log-odds.

\subsection{Results} 

Table~\ref{tab:cindex} reports the discriminative performance of ODDS-DR alongside baseline methods and ablation variants, evaluated using Harrell's C-index on the held-out validation sets. ODDS-DR matches or exceeds state-of-the-art survival models on both cohorts. On SUPPORT2, ODDS-DR paired with a log-logistic distribution achieves the highest overall discrimination, outperforming both deep survival baselines and linear approaches. On the CoA cohort---characterized by a low-sample regime and correspondingly wider confidence intervals---ODDS-DR's performance aligns closely with that of the robust linear Cox model, while retaining a tangible edge over more complex neural survival formulations.

\begin{table}[t!]
\centering
\small
\caption{{Discriminative performance (Harrell's C-index)} on the SUPPORT2 and CoA validation sets across different distribution choices for ODDS-DR, an ablation study using standard parameterizations (no reparameterization), and competing baseline models. For C-index computation, distributional models and DeepHit were ranked by the restricted area under the predicted survival curve; DeepSurv, CoxNAM, and Cox by the log-risk scores, and Cox-Time by the cumulative hazard evaluated at the median observed event time of the training set. Values in parentheses report 95\% confidence intervals obtained via non-parametric bootstrapping with 100 resamples. Bold denotes the best performance; underlined values indicate the second best.} \smallskip
\begin{tabular}{lccc}
\toprule
Model & Distribution & SUPPORT2 & CoA \\
\midrule

ODDS-DR (\textbf{ours}) & Gaussian
& 0.80 \scriptsize\color{gray}(0.79--0.81)
& \underline{0.66} \scriptsize\color{gray}(0.52--0.82)\\

ODDS-DR (\textbf{ours}) & Logistic
& 0.81 \scriptsize\color{gray}(0.80--0.82)
& 0.60 \scriptsize\color{gray}(0.44--0.77)\\

ODDS-DR (\textbf{ours}) & Log-logistic
& \textbf{0.83} \scriptsize\color{gray}(0.82--0.83)
& 0.64 \scriptsize\color{gray}(0.50--0.77)\\

\midrule

\textit{Ablation study:}\\

~~~no reparameterization & Gaussian
& 0.81 \scriptsize\color{gray} (0.80--0.82)
& 0.61 \scriptsize\color{gray}(0.44--0.77)\\

~~~no reparameterization & Logistic
& 0.80 \scriptsize\color{gray}(0.79--0.81)
& 0.61 \scriptsize\color{gray}(0.43--0.80)\\

~~~no reparameterization & Log-logistic
& \underline{0.82} \scriptsize\color{gray}(0.81--0.83)
& 0.57 \scriptsize\color{gray}(0.42--0.73)\\

\midrule

DeepSurv \cite{katzman2018deepsurv} & ---
& 0.81 \scriptsize\color{gray}(0.80--0.82)
& 0.59 \scriptsize\color{gray}(0.45--0.75)\\

Cox \cite{Cox1972} & ---
& 0.80 \scriptsize\color{gray}(0.79--0.81)
& \textbf{0.68} \scriptsize\color{gray}(0.54--0.81)\\

Cox-Time\slow{} \cite{KvammeBS19} & ---
& 0.81 \scriptsize\color{gray}(0.80--0.83)
& 0.58 \scriptsize\color{gray}(0.43--0.72)\\

DeepHit \cite{deephit} & ---
& 0.81 \scriptsize\color{gray}(0.80--0.82)
& 0.65 \scriptsize\color{gray}(0.51--0.79)\\

DSM \cite{nagpal2021deep} & ---
& \underline{0.82} \scriptsize\color{gray}(0.81--0.83)
& 0.62 \scriptsize\color{gray}(0.48--0.75)\\

CoxNAM \cite{XU2023120218} & ---
& 0.81 \scriptsize\color{gray}(0.80--0.82)
& 0.59 \scriptsize\color{gray}(0.37--0.81)\\

\bottomrule
\end{tabular}

\label{tab:cindex}
\end{table}

\smallskip

Importantly, the ablation study demonstrates that the reparameterization introduced in ODDS-DR---designed to enforce linear decodability of log-odds for instance-level interpretability---does not compromise discriminative accuracy. Across distribution choices, the log-odds-reparameterized variants consistently match or slightly exceed the accuracy of baseline parameterizations. We attribute this resilience and slight performance edge to the beneficial inductive bias imposed by our parameterization, which acts as a regularizing prior in regions of low data density and under incomplete feature observations (cf.\ \textbf{Property~1} introduced in Section~\ref{section:odds_dr}).

\subsection{Quantitative Evaluation of Extracted Risk Markers}

Beyond its time-to-event predictive accuracy, the primary methodological advantage of ODDS-DR lies in its structured design---namely, the linear decodability of log-odds---which facilitates the direct attribution of predicted outcomes to input clinical features. A key application of this functionality is clinical marker discovery. Because absolute ground-truth markers are rarely known in clinical cohorts, preventing direct validation, we resort to clinical guidelines and established observational studies as a proxy ground truth.

\smallskip

For the CoA cohort, reference features identified as clinically relevant for reintervention include the `pressure gradient across the coarctation', `body surface area (BSA)', and `age' \cite{esc_congenital, aha_aortic_disease, FIERLEY2026104041}. For the SUPPORT2 cohort, relevant predictors include `age', the `number of simultaneous diseases (comorbidities)', and validated clinical scoring systems, specifically a score based on the Glasgow Coma Scale \cite{support2_features_glasgow_scoma}, the physiology score \cite{support2_features_sps}, and the APACHE III score \cite{support2_features_apache}. For models that support a specific evaluation horizon (such as ODDS-DR and Cox-Time), horizons were set to 1 year for SUPPORT2 and 3 years for CoA, reflecting the time frames over which their respective clinical guidelines are most informative. We assess alignment with clinical guidelines via the Area Under the Receiver Operating Characteristic curve (AUROC), computed using absolute feature attributions averaged over the entire dataset. The results are summarized in Table~\ref{tab:attr_auc}.

\begin{table}[t!]
\centering \small
\caption{{Guideline alignment evaluated by Area Under the ROC Curve (AUROC)}, measuring how accurately feature attribution scores retrieve clinically established risk markers across the SUPPORT2 and CoA cohorts. Values in the right column denote 95\% confidence intervals computed via non-parametric bootstrapping. Bold indicates the highest AUROC per dataset. The hourglass symbol (\slow{}) marks attribution methods incurring substantial computational overhead. Layer-wise Relevance Propagation (LRP) is employed where analytically tractable; DeepHit is paired with Integrated Gradients (IG) \cite{SundararajanTY17} and SHAP \cite{Lundberg2020} due to softmax coupling; and Cox-Time's prediction being an integral creates a natural pairing with IG. DeepSurv, CoxNAM, and Cox regression are inherently time-invariant and denoted by `---'.} \smallskip
\label{tab:attr_auc}
  \makebox[\textwidth][c]{%
\begin{tabular}{llccccc}
\toprule
Dataset &
Method + Attribution &
Horizon &
\makecell{Guideline\\alignment\\(AUROC)} &
\makecell{95\% confidence\\interval}\\[1mm]
\midrule

\multirow{9}{*}{SUPPORT2}

& ODDS-DR (Log-logistic) + LRP (\textbf{ours}) & 1 year
& \textbf{0.89} & \em 0.88--0.90\\

& Cox-Time \cite{KvammeBS19} + IG\slow{} & 1 year
& 0.87 & \em 0.87--0.88\\

& DeepHit \cite{deephit} + IG\slow{} & 1 year
& 0.81 & \em 0.80--0.82\\

& DeepHit \cite{deephit} + SHAP\slow{} & 1 year
& 0.78 & \em 0.77--0.80\\

& DSM \cite{nagpal2021deep} + IG\slow{} & 1 year
& 0.84 & \em 0.83--0.85\\

& DSM \cite{nagpal2021deep} + SHAP\slow{} & 1 year
& 0.84 & \em 0.82--0.84\\

& DeepSurv \cite{katzman2018deepsurv} + LRP & --
& 0.86 & \em 0.84--0.88\\

& CoxNAM \cite{XU2023120218} + custom & --
& 0.76 & \em 0.76--0.78\\

& Cox + coefficient ranking & --
& 0.78\\

\midrule

\multirow{9}{*}{CoA}

& ODDS-DR (Log-logistic) + LRP (\textbf{ours}) & 3 years
& \textbf{0.84} & \em 0.73--0.88\\

& Cox-Time \cite{KvammeBS19} + IG\slow{} & 3 years
& 0.71 & \em 0.62--0.80\\

& DeepHit \cite{deephit} + IG\slow{} & 3 years
& 0.64 & \em 0.55--0.73\\

& DeepHit \cite{deephit} + SHAP\slow{} & 3 years
& 0.65 & \em 0.53--0.71 \\

& DSM \cite{nagpal2021deep} + IG\slow{} & 3 years
& 0.68 & \em 0.54--0.77\\

& DSM \cite{nagpal2021deep} + SHAP\slow{} & 3 years
& 0.66 & \em 0.56--0.72\\

& DeepSurv \cite{katzman2018deepsurv} + LRP & --
& 0.78 & \em 0.68--0.84\\

& CoxNAM \cite{XU2023120218} + custom & --
& 0.62 & \em 0.58--0.69\\

& Cox + coefficient ranking & --
& 0.66\\

\bottomrule
\end{tabular}
}

\end{table}

\smallskip

Across both datasets, ODDS-DR achieves strong alignment with clinical guidelines, yielding the highest AUROC scores on both benchmarks ($0.89$ on SUPPORT2 and $0.84$ on CoA), with a significant difference compared to simpler baselines such as Cox. While these results confirm the practical utility of our approach, we interpret them with appropriate nuance: clinical guidelines represent expert consensus rather than an absolute, exhaustive ground truth, and the wider confidence intervals on the smaller CoA cohort ($0.73$--$0.88$) reflect expected sampling variability.

\subsection{Qualitative Analysis}
\label{section:qualitative}

\begin{figure}[t!]
        \centering
          \makebox[\textwidth][c]{%
        \includegraphics[width=0.9\textwidth]{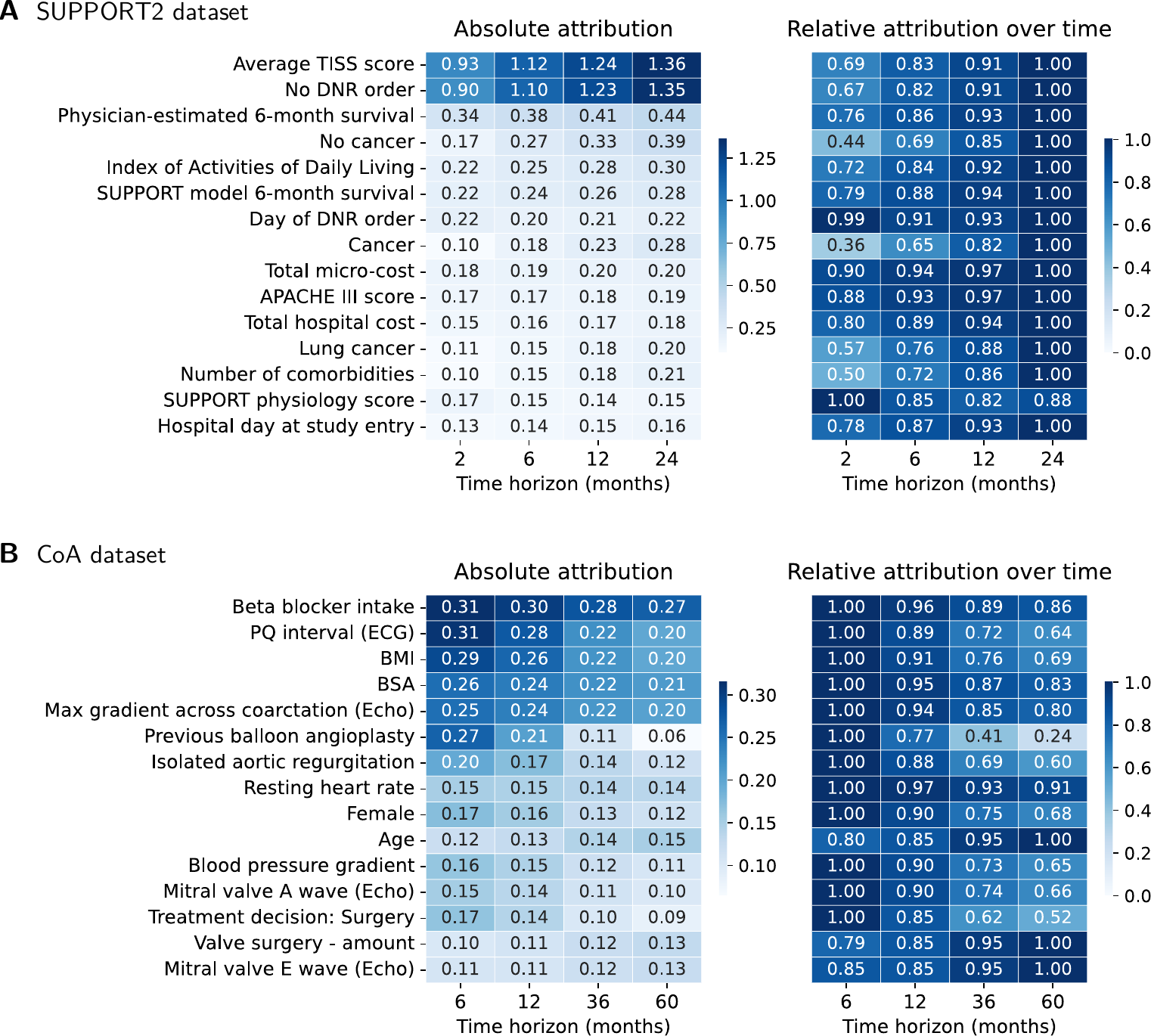}
        }
\caption{{Time-varying feature attribution across prediction horizons.} 
Results are shown for \textbf{(A)}~SUPPORT2 and \textbf{(B)}~CoA cohorts across increasing follow-up times. 
Denoting by $R_{it}^{(n)}$ the attribution of log-odds $L$ onto feature $x_i$ at horizon $t$ for patient $n$, the left panels display mean absolute attribution scores, $R_{it} = \mathbb{E}_n[|R_{it}^{(n)}|]$, reflecting overall feature importance. 
The right panels display feature-wise normalized scores, $R_{it} / \max_{t'} R_{it'}$, to isolate and compare relative temporal trajectories across clinical markers.}
\label{fig:heatmaps}
\end{figure}

While the quantitative evaluation above demonstrates competitive predictive performance and effective feature retrieval, the primary advantage of our approach over recent methods lies in its intrinsic interpretability, enabling the inference of clinical markers at the single-instance level robustly across multiple time horizons. We leverage this high analytic resolution to conduct an in-depth analysis of risk factors across both cohorts. In Fig.~\ref{fig:heatmaps}, we illustrate the dataset-wide importance of individual clinical features across several time horizons.

\smallskip

\smallskip

In the SUPPORT2 cohort, several clinical features emerge as primary markers of mortality risk. Specifically, TISS scores and a do-not-resuscitate (DNR) order drive risk and its progression across the entire time horizon. Other variables, such as a cancer diagnosis, contribute substantially to mortality but exhibit a distinct temporal dynamic: their impact is secondary at baseline but increases multifold after 24 months. In contrast, the predominantly short-term relevance of the TISS and physiology scores is clinically sound, as these instruments are purposely designed to index short-term mortality risk. 

\smallskip

In the CoA cohort, early risks are strongly attributable to specific clinical features, whereas long-term risk progression shows less concentrated attribution and remains partly unexplained. Among the primary drivers of early reintervention is `beta-blocker intake'---a marker that likely reflects underlying hypertension, which has been linked to an increased risk of early aortic re-coarctation \cite{FIERLEY2026104041}. An even more pronounced trend toward short-term reintervention is observed for `previous balloon angioplasty', where its role as a clinical marker attenuates sharply over longer horizons. A similar temporal pattern has been reported in previous cohorts, where the hazard of reintervention after balloon angioplasty was highest during early follow-up \cite{chiu2013coarctation}. This observation motivates the hypothesis that its prognostic value might be confined to the early period following the baseline visit, a pattern that remains to be confirmed in other cohorts.

\smallskip

The ability of our analysis to infer clinical variables that drive short- and long-term risk may prove useful as a stratification tool for clinical follow-up planning. For example, patients presenting with multiple early-risk factors could benefit from intensified monitoring and shorter surveillance intervals.

\section{Conclusion and Discussion}

We introduced ODDS-DR, an odds-informed parameterization of deep distributional regression that makes continuous probabilistic predictions directly amenable to feature attribution. By structuring network outputs to contribute linearly to event log-odds, the resulting explanations become continuous and parametric in time, cleanly disentangling short-term from long-term effects. Furthermore, rather than relying on costly model-agnostic explainers evaluated independently across discrete time horizons, ODDS-DR recovers entire feature-resolved risk trajectories in just two forward-backward passes, achieving orders-of-magnitude computational speedups.

\smallskip

Empirical results on real-world clinical time-to-event benchmarks corroborate the theoretical advantages of our method. Beyond matching the predictive performance of competitive proportional hazards and deep distributional baselines, ODDS-DR produces feature attributions that substantially align with clinical guidelines across their relevant time horizons, while resolving fine-grained temporal dynamics at the single-feature level.

\smallskip

Case studies on SUPPORT2 and the CoA cohort highlight distinct facets of clinical utility: on SUPPORT2, ODDS-DR scales computationally while delivering expressive, fully auditable nonlinear models; on the CoA cohort, it demonstrates high fidelity in a granular, single-center setting with dense multimodal monitoring---a regime representative of specialized clinical management for rare cardiovascular conditions, isolated from the cross-site batch effects that frequently confound multi-cohort evaluations.

\subsection*{Limitations and Future Work}

From a methodological standpoint, ODDS-DR currently relies on parametric families with closed-form log-odds expressions, confining the framework to unimodal density profiles. While unimodal formulations proved sufficient for our benchmark evaluations, capturing more complex hazard structures---such as those induced by competing events or non-susceptible subgroups---will demand richer, e.g., mixture-based representations. Because standard probability-space mixture models do not yield additive log-odds decompositions, enhancing expressiveness by designing mixture mechanisms directly within the log-odds domain represents a natural next methodological step.

\smallskip

From an application perspective, our clinical analysis is observational and subject to sample size constraints, particularly in the CoA cohort, which reflects an intrinsically small target population. Furthermore, our framework estimates prognostic risk under prevailing care patterns rather than evaluating counterfactual treatment regimes. Rather than supporting immediate therapeutic decisions, our findings yield exploratory hypotheses on how specific markers predict patient vulnerability under routine care.

\smallskip

Finally, the computational efficiency of ODDS-DR and its native compatibility with propagation-based attribution methods (such as Layer-wise Relevance Propagation) naturally invite extension to high-dimensional data modalities. In particular, applying ODDS-DR to baseline medical imaging to map time-resolved risk trajectories of lesion progression or structural tissue degeneration offers a promising avenue for future work.

\section*{Data Availability Statement}

The SUPPORT2 dataset \cite{knaus2020support} is publicly available and can be downloaded through the UCI Machine Learning Repository (\url{https://archive.ics.uci.edu/dataset/880/support2}). The CoA dataset \cite{FIERLEY2026104041} is subject to patient privacy and institutional data protection regulations. Data access requests must be directed to the corresponding author of the original study, subject to formal confidentiality agreements and required ethical approvals.

\section*{Ethics Statement}

The prospective observational study of patients with CoA has a registered protocol (\href{https://clinicaltrials.gov/study/NCT02591940}{NCT02591940}) and was approved by the institutional ethics committee (Charité–Universitätsmedizin Berlin: EA2/172/13) and conducted in accordance with the principles of the Declaration of Helsinki. All participants or their legal representatives provided written informed consent before enrolment.

\section*{Acknowledgements}

This work was funded by the German Ministry for Education and Research as BIFOLD\,--\,Berlin Institute for the Foundations of Learning and Data (ref.\ BIFOLD25B). G.M. is supported by the Einstein Center for Early Disease Interception (EC-EDI). M.K. has received funding from the Bundesministerium für Forschung, Technologie und Raumfahrt (BMFTR, Federal Ministry of Research, Technology and Space, Project: VADYS-ME, grant number: 01EJ2406A). M.K. and T.K. have received funding within the CHAIN project (Project ID: 101314833), supported by the European Union’s EU4Health programme.

\section*{Declaration of Generative AI and AI-assisted technologies in the writing process}

During the preparation of this work, the author(s) used Gemini 3.8 and Claude Opus 5.5 in order to improve formulations. After using this tool/service, the author(s) reviewed and edited the content as needed and take(s) full responsibility for the content of the publication.

\bibliographystyle{elsarticle-num}
\bibliography{references}

\appendix

\section{Technical Background} \label{appendix:derivations}

\subsection{Cox proportional hazards model}\label{appendix:cox}
The Cox proportional hazards model describes the instantaneous event rate, or hazard, conditional on a set of covariates $\bx$. The hazard function is defined as

\begin{equation}
h(t \mid \bx)
=
\lim_{\Delta t \rightarrow 0}
\frac{
P(t \leq T < t+\Delta t \mid T \geq t, \bx)
}{
\Delta t
},
\end{equation}
where $T$ denotes the event time. Equivalently, the hazard can be expressed in terms of the survival function $S(t \mid \bx)$ as

\begin{equation}
h(t \mid \bx)
=
-\frac{d}{dt}
\log S(t \mid \bx)
=
\frac{f(t \mid \bx)}{S(t \mid \bx)}
\end{equation}
where $f(t \mid \bx)$ denotes the corresponding probability density function. The Cox model assumes that the hazard can be factorized into a time-dependent baseline hazard $h_0(t)$ and a covariate-dependent relative risk term:

\begin{equation}
h(t \mid \bx)
=
h_0(t)
\exp\left(\boldsymbol{\beta}^{\top}\bx\right)
\end{equation}
where $\boldsymbol{\beta}$ denotes the vector of regression coefficients. The term $h_0(t)$ describes the baseline hazard and can be left unspecified. The proportional-hazards property follows directly by considering two individuals with covariates $\bx_i$ and $\bx_j$. Their hazard ratio is

\begin{equation}
\frac{
h(t \mid \bx_i)
}{
h(t \mid \bx_j)
}
=
\frac{
h_0(t)\exp(\boldsymbol{\beta}^{\top}\bx_i)
}{
h_0(t)\exp(\boldsymbol{\beta}^{\top}\bx_j)
}
=
\exp\left[
\boldsymbol{\beta}^{\top}
(\bx_i-\bx_j)
\right]
\end{equation}
Since the baseline hazard $h_0(t)$ cancels, the hazard ratio does not depend on time. Consequently, the classical Cox model assumes time-invariant relative effects of the covariates. A major advantage of the Cox model is that the regression coefficients can be estimated without explicitly specifying the baseline hazard. For an individual $i$ experiencing an event at time $t_i$, let

\begin{equation}
\mathcal{R}(t_i)
=
\{j : \tilde{T}_j \geq t_i\}
\end{equation}
denote the risk set containing all individuals who are still under observation and event-free immediately before $t_i$. Conditional on one event occurring at $t_i$, the contribution of individual $i$ to the partial likelihood is

\begin{equation}
\frac{
\exp(\boldsymbol{\beta}^{\top}\bx_i)
}{
\sum_{j \in \mathcal{R}(t_i)}
\exp(\boldsymbol{\beta}^{\top}\bx_j)
}
\end{equation}
The partial likelihood over all observed events is therefore

\begin{equation}
\mathcal{L}_{\mathrm{partial}}
=
\prod_{i:\delta_i=1}
\frac{
\exp(\boldsymbol{\beta}^{\top}\bx_i)
}{
\sum_{j \in \mathcal{R}(t_i)}
\exp(\boldsymbol{\beta}^{\top}\bx_j)
}
\end{equation}
where $\delta_i=1$ indicates an observed event. Taking the logarithm gives

\begin{equation}
\ell(\boldsymbol{\beta})
=
\sum_{i:\delta_i=1}
\bigg[
\boldsymbol{\beta}^{\top}\bx_i
-
\log
\Big(
\sum_{j \in \mathcal{R}(t_i)}
\exp(\boldsymbol{\beta}^{\top}\bx_j)
\Big)
\bigg]
\end{equation}
The regression coefficients are obtained by maximizing the partial log-likelihood.

\subsection{DeepSurv}

DeepSurv extends the Cox proportional hazards model by replacing the linear predictor
$\boldsymbol{\beta}^{\top}\bx$ with a nonlinear function
$g_{\boldsymbol{\theta}}(\bx)$ represented by a neural network. The hazard is consequently modeled as
\begin{equation}
h(t \mid \bx)
=
h_0(t)
\exp\left(
g_{\boldsymbol{\theta}}(\bx)
\right)
\end{equation}
The neural network is trained by minimizing the negative Cox partial log-likelihood,
\begin{equation}
\mathcal{J}_{\mathrm{Cox}}
=
-
\sum_{i:\delta_i=1}
\bigg[
g_{\boldsymbol{\theta}}(\bx_i)
-
\log
\Big(
\sum_{j \in \mathcal{R}(t_i)}
\exp\left(
g_{\boldsymbol{\theta}}(\bx_j)
\right)
\Big)
\bigg]
\end{equation}
DeepSurv therefore allows nonlinear relationships between the input variables and relative risk while retaining the proportional-hazards assumption. In particular, the relative effect of two individuals remains independent of time,
\begin{equation}
\frac{
h(t \mid \bx_i)
}{
h(t \mid \bx_j)
}
=
\exp\left[
g_{\boldsymbol{\theta}}(\bx_i)
-
g_{\boldsymbol{\theta}}(\bx_j)
\right]
\end{equation}
An important computational property of Cox partial likelihood is that the loss of an individual experiencing an event depends on all individuals belonging to its corresponding risk set. Consequently, observations are not independent within the loss function. The exact evaluation of the partial likelihood therefore requires access to the complete risk sets. Standard mini-batch training, in which subsets of observations are sampled independently, provides only an approximation to the full Cox objective. This contrasts with the distributional regression models considered in this work, for which the likelihood decomposes into independent per-sample contributions and can therefore be optimized accurately using mini-batches.

\subsection{Negative log-likelihood for right-censored distributional regression}\label{appendix:nll_censored}

Let $T$ denote the event time with probability density function
$f(t \mid \boldsymbol{\theta})$, cumulative distribution function
$F(t \mid \boldsymbol{\theta})$, and survival function
\begin{equation}
S(t \mid \boldsymbol{\theta})
=
1 - F(t \mid \boldsymbol{\theta})
\end{equation}
where the distribution parameters $\boldsymbol{\theta}$ are predicted by the neural network.
For each individual, we have
\begin{equation}
\tilde{T}_i
=
\min(T_i, C_i)
\end{equation}
where $C_i$ denotes the censoring time, together with the event indicator
\begin{equation}
\delta_i
=
\begin{cases}
1, & T_i \leq C_i \\
0, & T_i > C_i
\end{cases}
\end{equation}
For an observed event ($\delta_i = 1$), the likelihood contribution is given by the probability density evaluated at the observed event time,
\begin{equation}
\mathcal{L}_i
=
f(\tilde{T}_i \mid \boldsymbol{\theta}_i)
\end{equation}
For a right-censored observation ($\delta_i = 0$), the available information is that the event has not occurred up to the censoring time. Its likelihood contribution is therefore
\begin{equation}
\mathcal{L}_i
=
P(T_i > \tilde{T}_i \mid \boldsymbol{\theta}_i)
=
S(\tilde{T}_i \mid \boldsymbol{\theta}_i)
\end{equation}
Both cases can be written compactly as
\begin{equation}
\mathcal{L}_i
=
f(\tilde{T}_i \mid \boldsymbol{\theta}_i)^{\delta_i}
S(\tilde{T}_i \mid \boldsymbol{\theta}_i)^{1-\delta_i}
\end{equation}
The corresponding negative log-likelihood is
\begin{equation}
\mathcal{J}
=
-\sum_{i=1}^{N}
\left[
\delta_i
\log f(\tilde{T}_i \mid \boldsymbol{\theta}_i)
+
(1-\delta_i)
\log S(\tilde{T}_i \mid \boldsymbol{\theta}_i)
\right]
\end{equation}
Equivalently, using $S(t \mid \boldsymbol{\theta}) = 1 - F(t \mid \boldsymbol{\theta})$,
\begin{equation}
\mathcal{J}
=
-\sum_{i=1}^{N}
\left[
\delta_i
\log f(\tilde{T}_i \mid \boldsymbol{\theta}_i)
+
(1-\delta_i)
\log
\left(
1-F(\tilde{T}_i \mid \boldsymbol{\theta}_i)
\right)
\right]
\end{equation}
This formulation applies to any continuous parametric distribution for which the probability density and survival function can be evaluated (Weibull, Gaussian, logistic, log-logistic).

\section{Derivations of ODDS-DR Parameterizations}

\subsection{Log-odds for bell-shaped distributions}\label{appendix:log-odds-bell-shaped}
In this subsection, we provide derivations for the log-odds expressions of the two bell-shaped distributions used in the main paper, namely, the logistic and Gaussian distributions. In particular, we demonstrate that their log-odds can be expressed as $g(\frac{t-\mu}{\sigma})$, where $g$ is a monotonically increasing odd function, while $\mu$ and $\sigma$ are location and scale parameters of the distributions, respectively.

\paragraph{Log-odds for the Logistic distribution}
Let $T$ be a random variable following a logistic distribution with parameters $\mu$ and $\sigma > 0$. Its CDF is given by $F(t) = \frac{e^{\frac{t-\mu}{\sigma}}}{1 + e^{\frac{t-\mu}{\sigma}}}$. Hence, the log-odds are given by: 
\begin{align}
    L(t)
     = \log\left(\frac{P(T\leq t)}{P(T > t)}\right) &= \log\left(\frac{F(t)}{1-F(t)}\right) \\
    &= \log\left(\frac{e^{\frac{t-\mu}{\sigma}}}{1+e^{\frac{t-\mu}{\sigma}}} \cdot \frac{1}{\frac{1}{1+e^{\frac{t-\mu}{\sigma}}}}\right) = \frac{t - \mu}{\sigma}
\end{align}
We identify the desired structure by setting $g$ to the identity function.

\paragraph{Log-odds for the Gaussian distribution}

Let $T$ follow a Gaussian distribution with mean $\mu$ and variance $\sigma^2$. Its CDF is given by $F(t) = \frac{1}{2}\left(1 + \operatorname{erf}(\frac{t-\mu}{\sigma \sqrt{2}})\right)$, and the log-odds can be expressed as: 
\begin{align}
    L(t) &=  \log\left(\frac{F(t)}{1-F(t)}\right) = \log\left(\frac{1 + \operatorname{erf}\left(\frac{t-\mu}{\sigma \sqrt{2}} \right)}{1 - \operatorname{erf}\left(\frac{t-\mu}{\sigma  \sqrt{2}} \right)}\right)
     = 
    2\tanh^{-1}\left(\operatorname{erf}\left(\frac{t-\mu}{\sigma\sqrt{2}}\right)\right)
\end{align}
For $g(\xi) = 2 \tanh^{-1}(\operatorname{erf}(\xi/\sqrt{2}))$, we once again identify the desired structure.

\subsection{Log-odds for the Log-Logistic distribution}\label{appendix:log-odds-log-logistic}
In this subsection, we derive the log-odds expression for a random variable $T$ following the log-logistic distribution. Its CDF is given by $F(t) = \frac{t^{\beta}}{\alpha^{\beta} + t^{\beta}}$, where $\alpha$ and $\beta$ are the parameters of the distribution. The log-odds are given by:
\begin{align}
    L(t) &= \log\left(\frac{t^{\beta}}{\alpha^{\beta} + t^{\beta}} \cdot \frac{\alpha^{\beta} + t^{\beta}}{\alpha^{\beta}}\right) \\
    &= \beta \cdot (\log(t) - \log(\alpha)) \\
    &=
  {\underbrace{\beta \cdot (\log(t_0) - \log(\alpha))}_{\zloc}}
+
{\underbrace{\beta}_{\zshape}} \cdot (\log(t) - \log(t_0))
\end{align}
\end{document}